\documentclass{article}

\usepackage{microtype}
\usepackage{graphicx}
\usepackage{subcaption}
\usepackage{booktabs} 

\usepackage{hyperref}

\usepackage[accepted]{icml2026}

\usepackage{amsmath}
\usepackage{amssymb}
\usepackage{mathtools}
\usepackage{amsthm}

\usepackage[capitalize,noabbrev]{cleveref}

\theoremstyle{plain}

\theoremstyle{definition}

\theoremstyle{remark}

\usepackage[textsize=tiny]{todonotes}

\icmltitlerunning{Detecting and Mitigating Bias by Treating Fairness as a Symmetry Operation}

\begin{document}

\twocolumn[
  \icmltitle{Detecting and Mitigating Bias by Treating Fairness as a Symmetry Operation}



  \icmlsetsymbol{equal}{*}

  \begin{icmlauthorlist}
    \icmlauthor{Nishit Singh}{bits}
  \end{icmlauthorlist}

  \icmlaffiliation{bits}{Birla Institute of Technology and Science, Pilani}

  \icmlcorrespondingauthor{Nishit Singh}{f20221317@pilani.bits-pilani.ac.in}

  \icmlkeywords{Machine Learning, ICML}

  \vskip 0.3in
]



\printAffiliationsAndNotice{}  

\begin{abstract}
  Machine learning systems deployed in high stakes socioeconomic settings routinely display bias. We formalize bias as a symmetry breaking operation: a classifier is fair if its outputs remain invariant under the counterfactual operation of switching a sensitive attribute, with merit features held fixed. We implement loss based regularization as a symmetry restoring mechanism and evaluate the framework on four synthetic datasets with varying levels of noise, correlation, and bias. The framework achieves upwards of 90\% violation reduction, with accuracy costs around 5\%. This framework does not require causal graph knowledge, is computationally lightweight, and generalizes to any sensitive attribute definable as a bit-flip, making it suitable for contexts where local sources of discrimination remain absent from mainstream benchmarks.
\end{abstract}

\section{Introduction}

Machine Learning is increasingly being implemented in high-stakes decisions concerning real social groups. In these contexts, it is vital that the models and algorithms used are fair and just. However, such systems often exhibit systematic biases against socially sensitive groups, proving to be a significant source of harm \citep{hardt2016, chouldechova2017}. Previous research has proposed statistical definitions of fairness via criteria like demographic parity, equalized odds, and calibration as post-hoc constraints on a trained model. \citep{chouldechova2017, Kleinberg2016} prove that it is impossible to simultaneously satisfy calibration and equalized odds, opening the landscape for a fundamentally different approach: rather than imposing post-hoc constraints, we ask what structural invariance an unbiased model should satisfy by construction.

In this work, we treat fairness as a symmetry operation, and bias as a symmetry breaking operation. Borrowing the language of group theory and physics, a system has a symmetry if it is invariant under a group action. Formally, a classifier $f$ is invariant under action $T$ if $f(x) = f(Tx)$ for all $x \in \mathcal{X}$. In our setting, $T$ is the counterfactual operator that flips all sensitive attributes\footnote{Attributes like caste, gender, race etc.} while holding merit features\footnote{Attributes like education, experience, age etc.} fixed $T(\mathbf{x}) = [\mathbf{x}_m;\ \mathbf{1} - \mathbf{x}_s]$ where $\mathbf{x}_m$ are merit features and $\mathbf{x}_s$ are sensitive  attributes. This line of work closely follows the paradigm of symmetry-aware ML in texts like \citet{cohen2016}. This invariance condition is an observational flip, and thus requires no causal graph knowledge, making it simpler and more tractable than counterfactual fairness \citep{kusner2017}, at the cost of not accounting for causal mediation through correlated features.

\section{Problem Formulation}

\subsection{Setup}

We define a probabilistic classifier $f : X \to [0,1]$. Under the model, $f(x) = P(y=1| x)$, where $y$ is a binary label $y = \{0, 1\}$, and $x \in X$. $x$ is partitioned as $x = [x_m;x_s]$, where $x_m$ are merit features and $x_s$ are sensitive attributes. We define the transformation $T : X \to X$ as $T(x) = [x_m; 1-x_s]$.

\subsection{Bias as Symmetry Breaking}

A classifier is symmetric; or $T$-invariant if $f(x) = f(Tx)$ for all $x \in X$. Similar to \citet{dwork2012}, we define pointwise violation as $v(x) = |f(Tx) - f(x)|$, and population violation as $V = E_{x \sim P_X} [ |f(x) - f(T(x))| ]$, where $P(X)$ is the data distribution. Practically, population violation is approximated by the empirical mean $\hat{V} = \frac{1}{n} \sum_{i = 1}^{n} |f(Tx_i) - f(x_i)|$.

\begin{table*}[t]
\centering
\caption{Dataset taxonomy where $\gamma = [\gamma_{gender}, \gamma_{race}]$ and $d = 6$ (2 sensitive attributes + 4 merit features).}
\label{tab:datasets}
\renewcommand{\arraystretch}{1.3}
\begin{tabular}{lcccccc}
\toprule
\textbf{Dataset} & \textbf{$d$} & \textbf{$\boldsymbol{\gamma}$} & 
\textbf{Feat.--Sens.} & \textbf{Class} & \textbf{Noise} & 
\textbf{Pos.} \\
 & & & \textbf{Correlation} & \textbf{Imbalance} & 
\textbf{Distractors} & \textbf{Rate} \\
\midrule
$D_1$: Low-bias    & 6  & $[0.50,\ 0.38]$ & \texttimes & \texttimes & \texttimes & ${\sim}47\%$ \\
$D_2$: Low-bias and correlated  & 6  & $[0.50,\ 0.38]$ & \checkmark & \texttimes & \texttimes & ${\sim}50\%$ \\
$D_3$: High-bias and imbalanced & 6  & $[1.80,\ 1.35]$ & \texttimes & \checkmark & \texttimes & ${\sim}15\%$ \\
$D_4$: High-bias and noisy   & 12 & $[1.80,\ 1.35]$ & \checkmark & \texttimes & \checkmark & ${\sim}47\%$ \\
\bottomrule
\end{tabular}
\end{table*}

\subsection{Dataset Generation}

The label $y$ is drawn from
\[
    y \mid x \sim \text{Bernoulli}\!\left(\sigma\!\left(\beta_0 + \boldsymbol{\beta}^\top x_m + \boldsymbol{\gamma}^\top x_s\right)\right)
\]

\noindent where $\sigma(z) = (1 + e^{-z})^{-1}$ is the logistic sigmoid, $x_m = [\texttt{age},\ \texttt{years\_exp},\ \texttt{education},\ \texttt{skill\_score}]^\top$ are the merit features, $x_s = [\texttt{gender},\ \texttt{race}]^\top$ are the sensitive attributes, $\boldsymbol{\beta} = [0.0,\ 0.04,\ 0.70,\ 0.035]^\top$ are the merit coefficients, $\beta_0 = -3.0$ is the intercept, and $\boldsymbol{\gamma}$ is the injected bias vector whose values vary by dataset as described in Table 1.

The baseline employment probability (average merit, no bias) is about 4.7\%, reflecting competitive hiring. Education is given the highest influence on merit, followed by experience and skill. The symmetry breaking term is the bias injection, which is [0.5, 0.375] for low bias datasets and [1.8, 1.35] for high bias datasets. If $\boldsymbol{\gamma} = 0$, $V = 0$ by construction.

In order to stress the model, we also generate a correlation between merit features by generating them as a function of sensitive features\footnote{Through the correlation injection $\text{edu}_i \leftarrow \text{clip(edu}_i + 0.6 \cdot \text{gender}_i + \epsilon_i,  0, 3),   \epsilon_i \sim N(0, 0.09)
\textbf{ and }\text{skill}_i \leftarrow \text{clip(skill}_i + 8 \cdot \text{gender}_i + \delta_i,  0, 100),   \delta_i \sim N(0, 9)$}. We also introduce noise\footnote{The noise is generated by $n_k^{(i)} = \epsilon_k^{(i)} + \delta_k \cdot s^{(i)}$, where $n_k^{(i)}$ is the $k$th noise feature for sample $i$, $\epsilon_k^{(i)} \sim \mathcal{N}(0, 1)$, $s_i \in \{0,1\}$ is the sensitive attribute, and $\delta_k$ is the spurious correlation coefficient.} in a high bias dataset, which are 6 additional features which carry no meaningful signal about $y$, and made to have a small spurious correlation with the sensitive attributes.

\subsection{Loss Based Regularization}
The full objective is:
\[
    \mathcal{L}(\mathbf{w},b) = \mathcal{L}_{task}(\mathbf{w}, b) + \lambda\mathcal{L}_{sym}(\mathbf{w},b)
\]
Where $\mathcal{L}_{task}$ is the standard task loss by binary cross entropy:
\[
    \mathcal{L}_{task} = -\frac{1}{n} \sum_{i = 1}^{n}[y_ilog\space f(x_i) + (1-y_i)log \space (1-f(x_i))]  
\]
And $f(x) = \sigma(\mathbf{w}\top \tilde{x} + b); \space \sigma(z) = 1/(1+e^{-z})$. We define symmetric loss $\mathcal{L}_{sym}$ as: 
\[ \mathcal{L}_{sym} = \frac{1}{n} \sum_{i = 1}^{n}[f(x_i) - f(Tx_i)]^2\]
Now, we define the gradient for the new loss function $\mathcal{L}$ by defining the prediction gap for sample $i$ as $\Delta_i = f(\mathbf{x}_i) - f(T(\mathbf{x}_i))$. Applying the chain rule
to $\mathcal{L}_{\text{sym}} = \frac{1}{n}\sum_{i=1}^{n} \Delta_i^2$:
\[
    \frac{\partial \mathcal{L}_{\text{sym}}}{\partial \mathbf{w}} 
    = \frac{2}{n} \sum_{i=1}^{n} \Delta_i \cdot \frac{\partial \Delta_i}{\partial \mathbf{w}}
\]
Since $f(\mathbf{x}) = \sigma(\mathbf{w}^\top \tilde{\mathbf{x}} + b)$ and $\sigma'(z) = \sigma(z)(1-\sigma(z))$:
\[
    \begin{split}
    \frac{\partial \Delta_i}{\partial \mathbf{w}} ={} & f(\mathbf{x}_i)\bigl(1-f(\mathbf{x}_i)\bigr)\tilde{\mathbf{x}}_i \\
    & - f(T(\mathbf{x}_i))\bigl(1-f(T(\mathbf{x}_i))\bigr)\widetilde{T(\mathbf{x}_i)}
\end{split}
\]
Substituting:
\[
    \begin{split}
    \frac{\partial \mathcal{L}_{\text{sym}}}{\partial \mathbf{w}} = \frac{2}{n} \sum_{i=1}^{n} \Delta_i \Bigl[ & f(\mathbf{x}_i)\bigl(1-f(\mathbf{x}_i)\bigr)\tilde{\mathbf{x}}_i \\
    & - f(T(\mathbf{x}_i))\bigl(1-f(T(\mathbf{x}_i))\bigr)\widetilde{T(\mathbf{x}_i)} \Bigr]
\end{split}
\]
The update rule for learning rate $\eta$ with the new full objective gradient is 
\[
\mathbf{w} \leftarrow \mathbf{w} - \eta \left( \frac{\partial \mathcal{L}_{\text{task}}}{\partial \mathbf{w}} + \lambda \frac{\partial \mathcal{L}_{\text{sym}}}{\partial \mathbf{w}} \right)
\]
\section{Experiment}
We evaluate the loss regularization method across four synthetic datasets of increasing structural complexity. All experiments use n=2000 samples, a 75/25 train/test split stratified by label where the positive rate exceeds 5\%. The violation metric $V$ is computed on both train and test sets using soft predicted probabilities. For the loss regularization method we sweep $\lambda \in \{0.0, 0.5, 1.0, 2.0, 5.0, 10.0\}$, where $\lambda = 0$ recovers the unregularized baseline. For dataset $D_2$ (low bias and correlated), we find the predictive accuracies to be comparable, as shown in Figure 1. Comparing the violation metric, we see a stark 93.2\% decrease in the violation on the test set for $D_2$, as visualized in Figures 2 and 3. Accuracies and violations are visualized against the $\lambda$ sweep for all datasets in Figure 4. Comparisons for violation between the baseline model and regularized model for all datasets can be found in the Appendix. 
\begin{figure}[ht]
  \centering
  \includegraphics[width=\columnwidth]{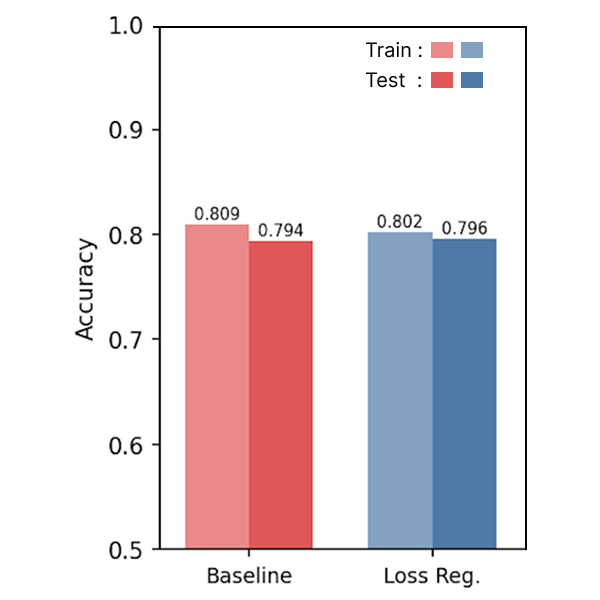}
  \vspace{-1mm}
  \caption{Comparison of accuracies of the baseline model vs. the regularized model of the dataset $D_2$.}
  \label{fig:violations} 
  \vspace{-3mm}
\end{figure}

\begin{figure}[ht]
  \vskip 0.2in
  \begin{center}
    \centerline{\includegraphics[width=\columnwidth]{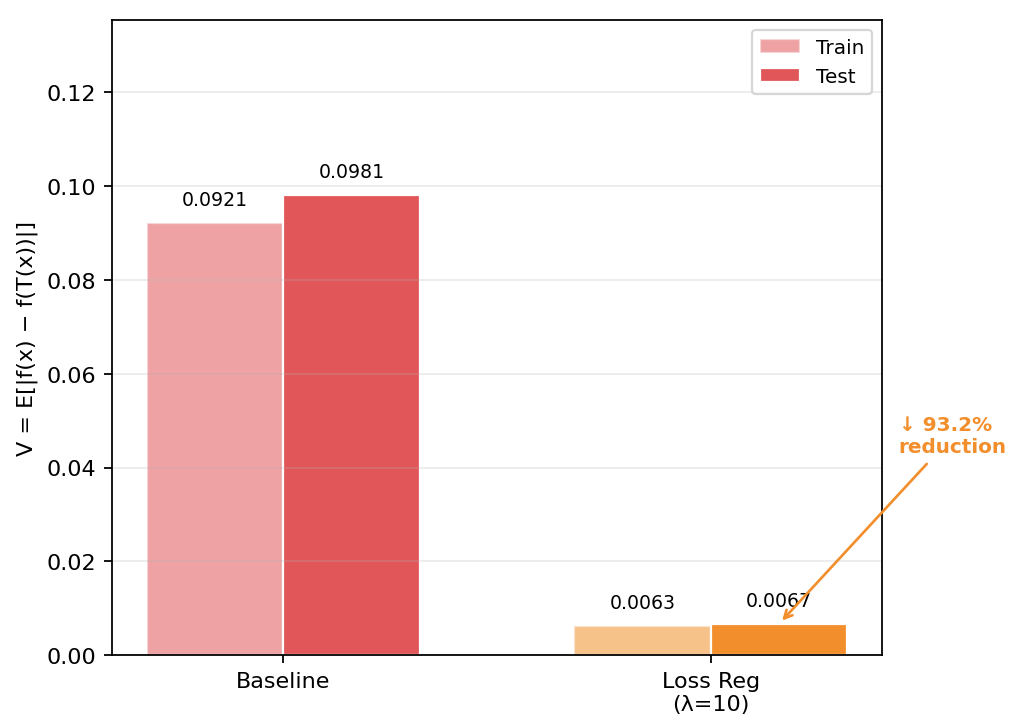}}
    \caption{
      Comparison of the violation by the baseline model vs. the regularized model.
    }
    \label{icml-historical}
  \end{center}
\end{figure}

\begin{figure}[ht]
  \vskip 0.2in
  \begin{center}
    \centerline{\includegraphics[width=\columnwidth]{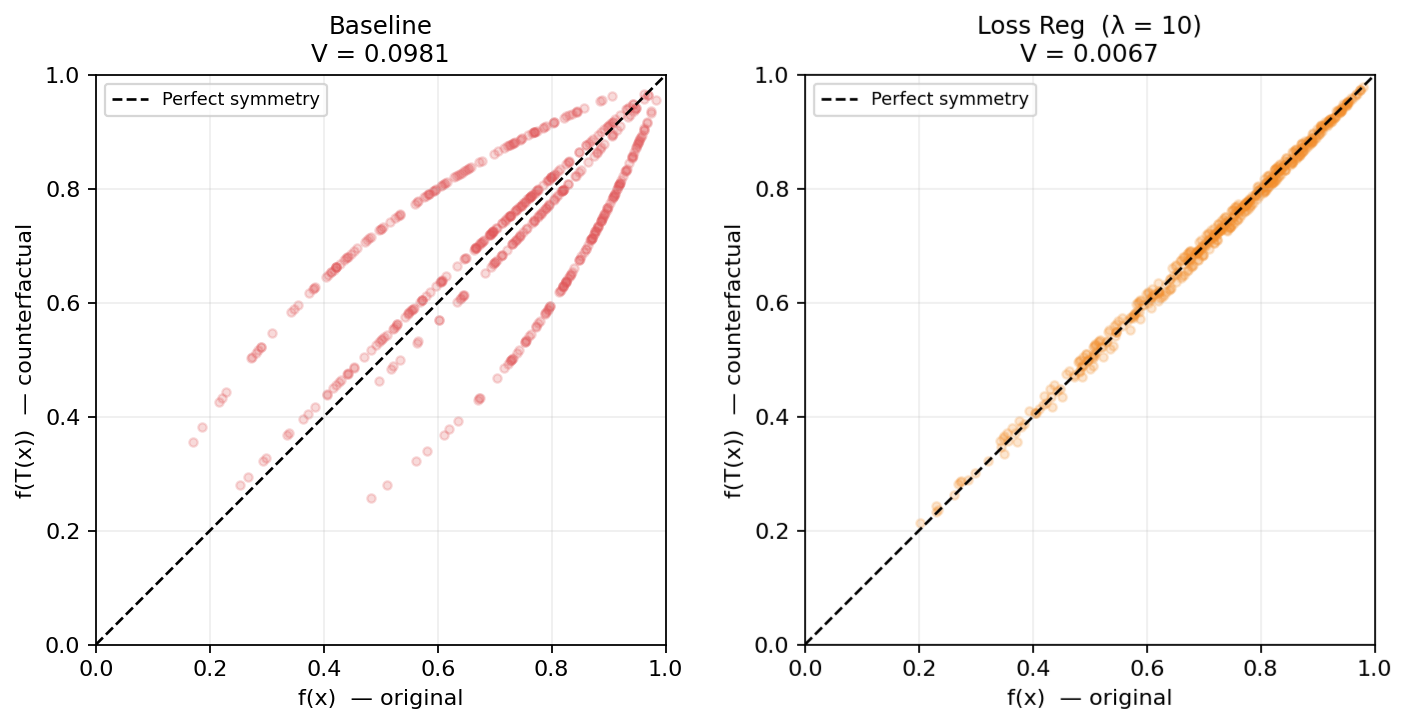}}
    \caption{
      Scatter plots of the outputs of the baseline model vs. the regularized model.
    }
    \label{icml-historical}
  \end{center}
\end{figure}

\begin{figure*}[ht]
  \vskip 0.2in
  \begin{center}
    \centerline{\includegraphics[width=\textwidth]{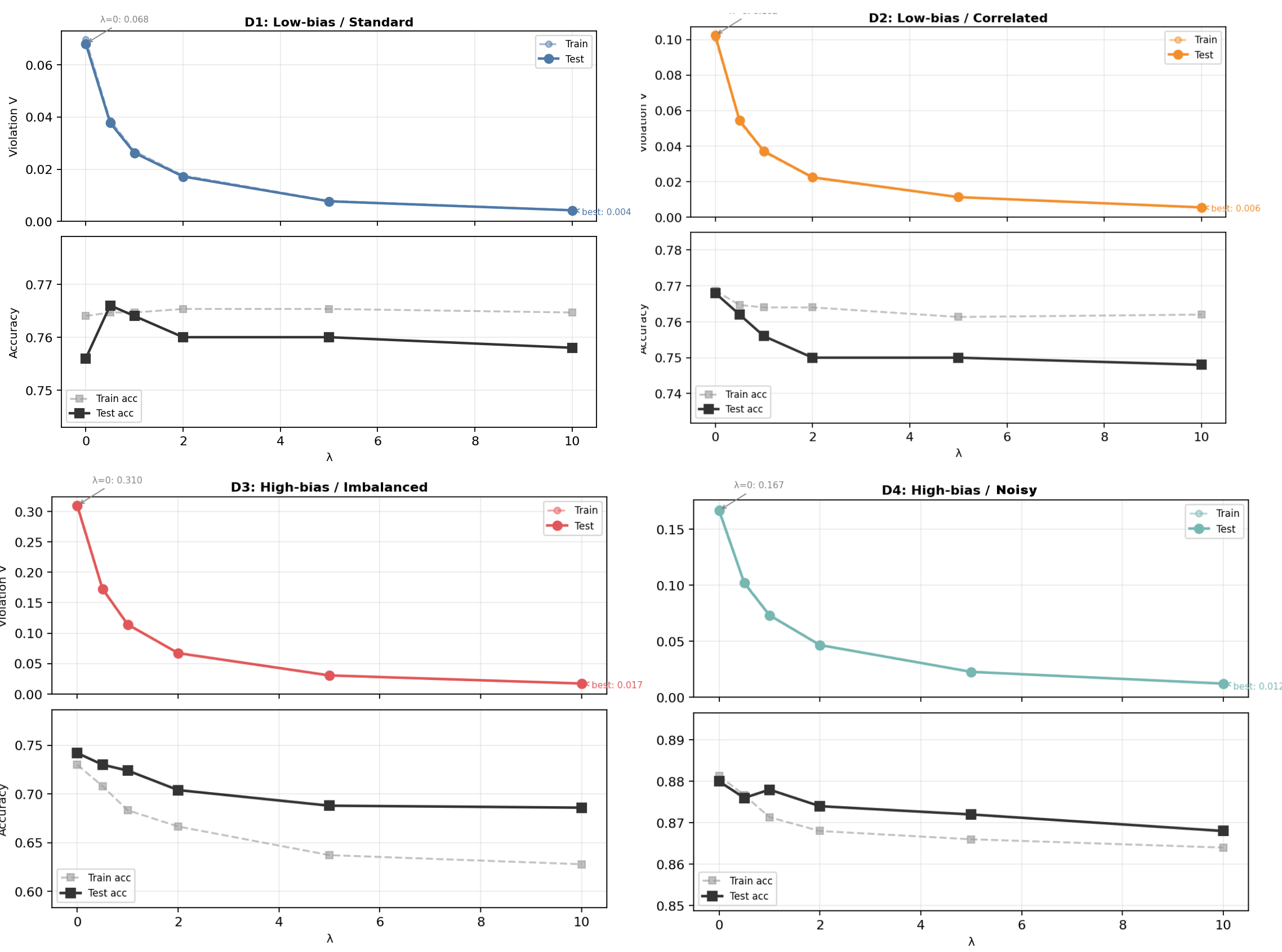}}
    \caption{
      $\lambda$ vs. Violation and $\lambda$ vs. Accuracy for all datasets $D_1$, $D_2$, $D_3$ and $D_4$.
    }
    \label{icml-historical}
  \end{center}
\end{figure*}

\section*{Related Work}

Algorithmic bias has extensively been studied through the lens of statistical criteria as applied post-hoc constraints. \citet{hardt2016} introduced equalized odds, requiring equal true and false positive rates across demographics. \citet{chouldechova2017} and \citet{Kleinberg2016} independently prove that calibration and equalized odds cannot simultaneously hold when base rates differ across groups, motivating our approach to study the structural invariance of an unbiased model. 

In-processing methods, like the models proposed by \citet{kamishima2012} add a prejudice regularizer\footnote{The mutual information between prediction and sensitive attributes.} to logistic regression during training. \citet{zhang2018} takes an adversarial approach by training an adversary to retrieve the sensitive attributes from the predictions of the model. 

Our work is most closely related to counterfactual fairness \citep{kusner2017}, which defines a predictor as unbiased if the output is invariant to interventions on sensitive attributes in a structural causal model. \citet{kilbertus2017} extend this line of work by blocking discriminatory pathways identified via causal path analysis. Our work is a deliberate simplification in order to be functional in contexts without a causal graph, with the tradeoff of not accounting for causal mediation through correlated features, stress-tested in dataset $D_2$.

\section*{Discussion}
Automated decision making systems are being deployed rapidly all over the world to aid in critical processes like hiring, welfare allocation, and financial analysis \citep{okolo2020}. These deployments often have less regulatory oversight on the demographics which are underrepresented in the datasets used to develop them \citep{joseph2025}, motivating research in fairness and bias mitigation methods. 

In this work, we have utilized synthetic data with built in structural bias as a low resource tool. Popular fairness benchmarks\footnote{COMPAS, Adult Income, etc.} are predominantly western in origin and do not effectively encode a lot of prejudices and biases from countries from the Global South \citep{sambasivan2021}. The symmetry violation and regularizer were developed, validated and researched on datasets with a known data generation process, providing a framework for fairness research in data scarce environments.

The data generation process is also general by design. The framework does not require a redesign to include different protected groups, since the biases encoded in $x_s$ can be redefined and the coefficients readjusted to reflect the diverse socioeconomic landscapes of different regions, proving to be effective in modeling different industries with different sensitive attributes like healthcare, hiring, and lending\footnote{Attributes like age might be a sensitive factor in healthcare and lending and not hiring; similarly education might be a sensitive factor for hiring and lending, not healthcare.}.

This bias mitigation paradigm is also computationally lightweight. Since this design does not require causal graph knowledge or training an adversary to identify and mitigate bias, the model at cost of causal negligence provides a significant compute discount.

\nocite{langley00}

\bibliography{example_paper}
\bibliographystyle{icml2026}

\newpage
\appendix
\onecolumn
\section{Appendix}

\subsection{Bar Charts for violation comparison between the baseline model and loss regularized model}
\begin{figure}[ht]
  \vskip 0.2in
  \centering
    \centerline{\includegraphics[width=\textwidth]{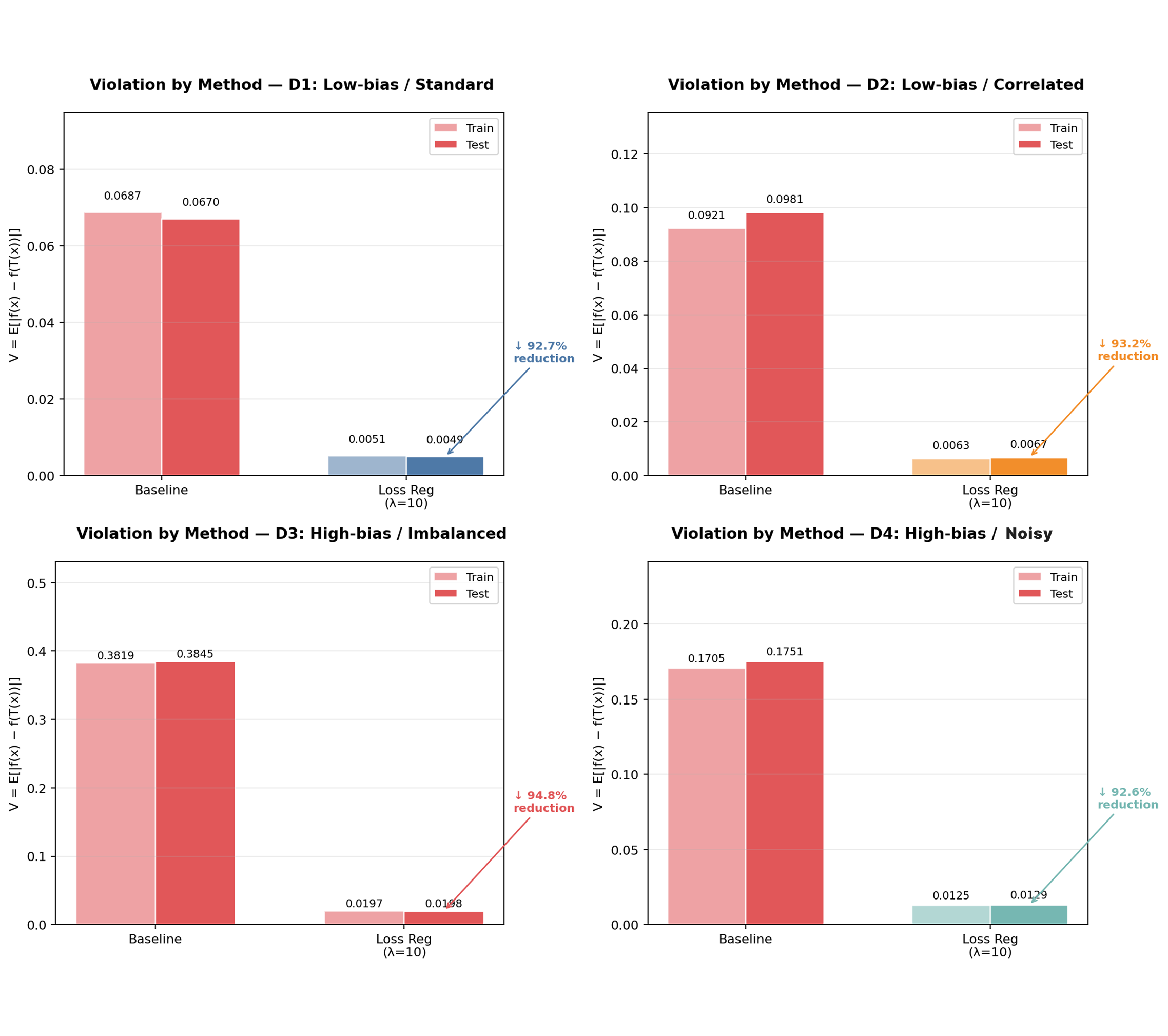}}
    \caption{
      Loss regularized models vs. baseline model.
    }
    \label{icml-historical}
  
\end{figure}
\newpage
\subsection{Scatter Plots for Violation Comparison Bet}

\begin{figure}[ht]
  \vskip 0.2in
  \begin{center}
    \centerline{\includegraphics[width=\textwidth]{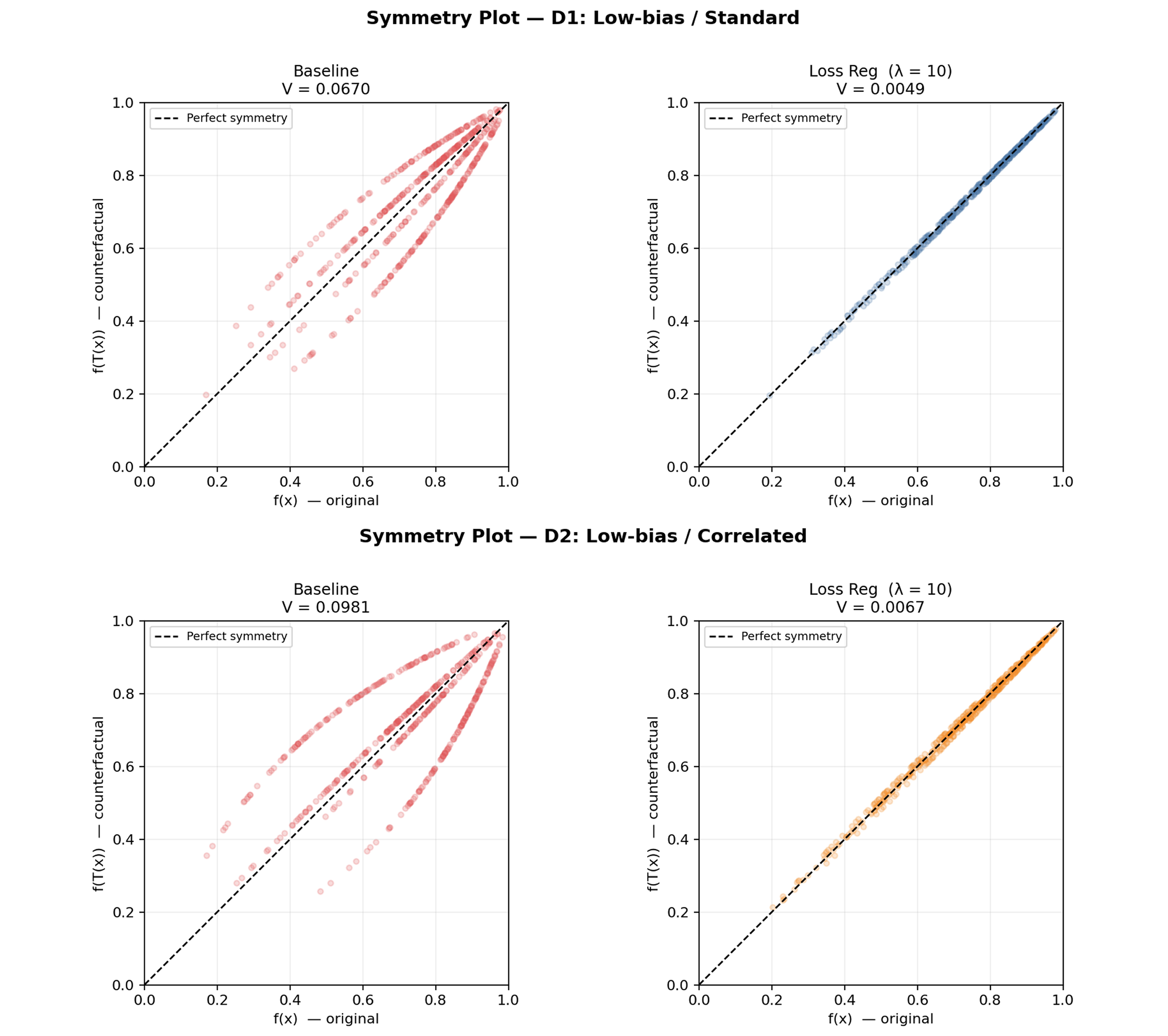}}
    \caption{
      Loss regularized models (D1, D2) vs. baseline model.
    }
    \label{icml-historical}
  \end{center}
\end{figure}

\begin{figure}[ht]
  \vskip 0.2in
  \begin{center}
    \centerline{\includegraphics[width=\textwidth]{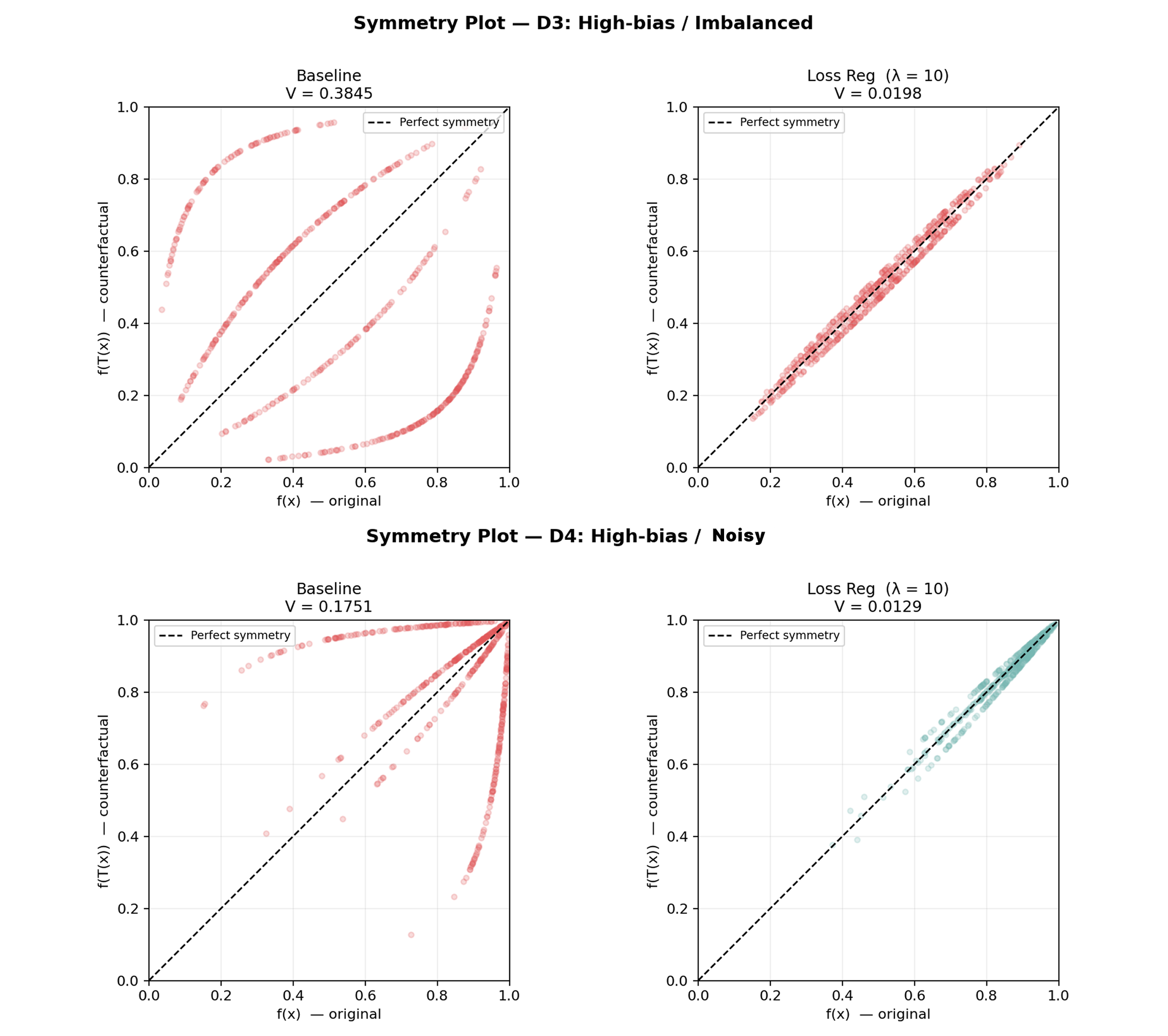}}
    \caption{
      Loss regularized models (D3, D4) vs. baseline model.
    }
    \label{icml-historical}
  \end{center}
\end{figure}
\end{document}